# Trajectory-Aware Reliability Modeling of Democratic Systems

Dmitry ZAYTSEV[1], Valentina KUSKOVA[2], Michael COPPEDGE[2]

[1]Lucy Family Institute for Data & Society, University of Notre Dame, USA
[2]Department of Political Science, University of Notre Dame, USA

**Keywords:** Reliability modeling; Degradation propagation; Failure risk prediction; Dynamic causal networks; Neural time-series models; Democratic system stability

**Abstract.** Failures in complex systems often emerge through gradual degradation and the propagation of stress across interacting components rather than through isolated shocks. Democratic systems exhibit similar dynamics, where weakening institutions can trigger cascading deterioration in related institutional structures. Traditional reliability and survival models typically estimate failure risk based on the current system state but do not explicitly capture how degradation propagates through institutional networks over time. This paper introduces a trajectory-aware reliability modeling framework based on Dynamic Causal Neural Autoregression (DCNAR). The framework first estimates a causal interaction structure among institutional indicators and then models their joint temporal evolution to generate forward trajectories of system states. Failure risk is defined as the probability that predicted trajectories cross predefined degradation thresholds within a fixed horizon. Using longitudinal institutional indicators, we compare DCNAR-based trajectory risk models with discrete-time hazard and Cox proportional hazards models. Results show that trajectory-aware modeling consistently outperforms Cox models and improves risk prediction for several propagation-driven institutional failures. These findings highlight the importance of modeling dynamic system interactions for reliability analysis and early detection of systemic degradation.

## 1. Introduction

Complex systems rarely fail suddenly. In many domains, system failures emerge through gradual degradation processes that propagate across interacting subsystems rather than through isolated component breakdowns [1]. This pattern is widely observed in engineering infrastructures, where wear or malfunction in one subsystem can increase stress on others and trigger cascading failures. Similar dynamics arise in financial networks, where shocks to one institution propagate through interconnected balance sheets and liquidity channels. These examples illustrate a broader principle: reliability in complex systems depends not only on the state of individual components, but also on the structure of interactions among them [2].

Socio-institutional systems exhibit analogous dynamics. Democratic governance, for example, is sustained by multiple interacting institutional components, including electoral processes, civil liberties, legislative constraints, and citizen participation [3]. Weakening in one institutional domain can propagate stress to others, producing systemic degradation over time. Declines in civil liberties may undermine electoral accountability; weakening electoral integrity may reduce constraints on executive authority; and deteriorating institutional checks can further accelerate democratic erosion. From a reliability perspective, such dynamics resemble cascading degradation processes in engineered systems, where component interactions shape the overall stability of the system [4].

Traditional reliability and survival models provide useful tools for estimating system failure risk, but they typically rely on simplifying assumptions that may limit their applicability in complex institutional environments [4]. Many reliability approaches assume independent component failures or static dependency structures, while survival models generally estimate failure risk as a function of the current system state. In these formulations, risk is often modeled as $P(F_t|X_t)$, where failure probability depends on the present configuration of system indicators. Although such models can effectively capture threshold-driven risks, they do not explicitly represent how degradation propagates through interacting subsystems over time. As a result, they may fail to detect failures that emerge from dynamic interactions among system components.

This paper introduces a trajectory-aware reliability modeling framework designed to capture degradation propagation in complex institutional systems. The proposed approach is based on Dynamic Causal Neural Autoregression (DCNAR) [5], which integrates causal network discovery with nonlinear time-series modeling. The framework first estimates a causal interaction structure among institutional indicators, representing pathways through which degradation can propagate. It then models the joint temporal evolution of these indicators to generate forward trajectories of system states [16]. Failure risk is estimated from these predicted trajectories, allowing the model to identify cases in which future degradation paths are likely to cross critical thresholds.

Empirically, we evaluate the proposed framework using longitudinal institutional indicators that characterize multiple components of democratic governance. We compare DCNAR-based trajectory risk models with classical survival approaches, including discrete-time hazard models [6] and Cox proportional hazards models [7]. The results show that trajectory-aware modeling consistently outperforms Cox models across most institutional indicators and improves risk prediction relative to classical hazard models for several propagation-driven failures. These findings suggest that dynamic interaction structures play a critical role in shaping institutional reliability and highlight the importance of trajectory-based approaches for modeling failure risk in complex socio-institutional systems.

## 2. Reliability Modeling of Democratic Systems

Reliability engineering provides a useful conceptual framework for analyzing the stability of complex systems composed of interacting components. In traditional engineering applications, system reliability refers to the probability that a system continues to perform its intended function over time despite degradation in individual components [8]. The same conceptual framework can be applied to socio-institutional systems, where the stability of democratic governance depends on the functioning of multiple institutional subsystems.

**System Representation.** Let the state of a democratic system at time $t$ be represented by a vector of institutional indicators $X_t = (x_{1t}, x_{2t,} \dots, x_{nt})$, where each component $x_{it}$ represents the observed level of a specific institutional attribute at time *t.* These attributes correspond to measurable indicators of institutional performance, such as electoral integrity, freedom of expression, civil participation, or legislative constraints on executive authority. Together, these indicators describe the operational condition of the democratic system.

In this representation, each indicator can be interpreted as a component health measure, analogous to the state of a subsystem in engineering reliability analysis. Higher values generally correspond to stronger institutional performance, while declining values reflect institutional degradation. System reliability can be defined as the probability that the system remains within a region of institutional stability over time.

**System Reliability.** Let $\mathcal{H} \subset \mathbb{R}^n$denote the set of system states corresponding to stable democratic functioning. The reliability of the system at time $t$ can then be expressed as $R(t) = P(x_t \in \mathcal{H})$, which represents the probability that the institutional configuration at time $t$ remains within the stability region. In practice, the stability region is defined operationally through thresholds on institutional indicators. When key institutional components deteriorate beyond critical levels, the system enters a degraded state that increases the likelihood of systemic instability.

**Failure Events.** A system failure occurs when the system state leaves the stability region. Formally, we define a failure indicator $F_t = 1(X_t \notin \mathcal{H})$, which takes value 1 when the system is in a degraded state and 0 otherwise. Operationally, failure events are identified by threshold conditions on institutional indicators. For example, if an indicator measuring electoral integrity or civil liberties falls below a predefined threshold, the system is considered to have entered a degraded reliability state. These thresholds can be determined empirically, such as by defining failure as values below a specified quantile of the historical distribution.

**Degradation Indicators**. Institutional indicators serve as degradation signals that reflect the health of the democratic system. Examples include measures of electoral fairness, freedom of expression, judicial independence, and political participation. Over time, gradual deterioration in these indicators signals increasing stress within the system. Unlike many engineering systems in which component degradation may be independent, institutional indicators are often highly interdependent [9]. Declines in one domain may influence other domains, producing correlated degradation patterns across the system. For example, erosion of judicial independence may weaken constraints on executive power, which in turn can affect electoral competitiveness or civil liberties.

From a reliability perspective, these indicators collectively describe the evolving condition of the system. Monitoring their trajectories provides information about both current system's health and the potential for future failures.

## 3. Degradation Propagation in Institutional Systems

In many complex systems, failures do not arise from isolated component breakdowns but from the propagation of degradation across interacting subsystems. Reliability engineering has long recognized that the behavior of complex systems depends not only on the health of individual components but also on the structure of their interactions [10]. When components are interconnected, deterioration in one subsystem can propagate through the network and amplify system-level risk.

Institutional systems exhibit similar dynamics. Democratic governance relies on multiple institutional components that interact through formal rules, legal constraints, and political practices. When one institutional domain deteriorates, it can increase stress on others, potentially initiating a cascading process of institutional degradation. For example, weakening judicial independence may reduce the ability of courts to enforce constitutional constraints. This can allow executive actors to circumvent legal limits on power, which in turn may affect electoral competitiveness or restrict media freedoms. Similarly, declines in civil liberties can reduce citizen participation and weaken mechanisms of political accountability, propagating through the institutional system and increasing the risk of democratic breakdown (Figure 1).

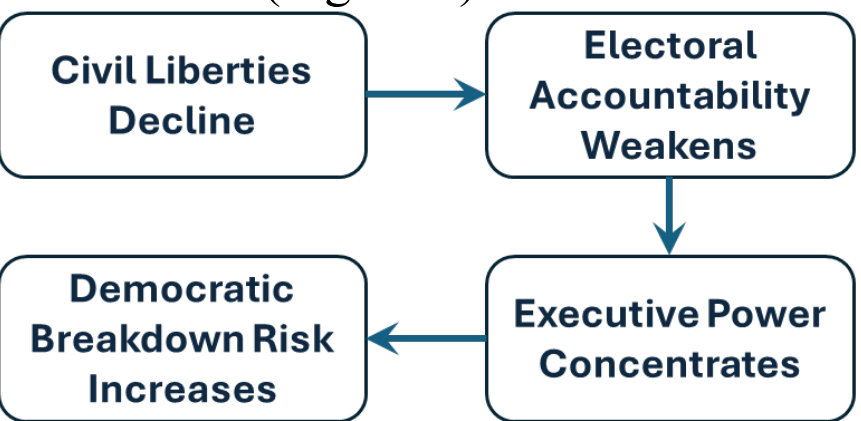

**Figure 1. Degradation Propagation in Institutional Systems Example**

From a reliability perspective, such dynamics can be represented as a network of interacting components. Let $A = (a_{ij})$ denote the matrix describing directional relationships among institutional indicators, where $a_{ij}$ represents the influence of component $j$ on component $i$. In this formulation, edges in the network capture pathways through which degradation can propagate. A decline in component $j$ may increase the probability that component $i$ will deteriorate in subsequent periods.

This perspective highlights an important limitation of traditional survival models used in political and social analysis. Standard hazard models typically estimate failure risk as a function of the current system state, often expressed as $P(F_t \mid X_t)$. While this formulation captures threshold-driven risks, it does not explicitly model how degradation evolves over time through interactions among components. In systems where failures emerge through dynamic propagation, the current state alone may not fully capture the risk trajectory of the system.

Instead, failure risk may depend on the future trajectory of institutional indicators, which is shaped by both their current values and the interaction structure among them. In such cases, the relevant quantity is not only the probability of immediate failure but also the likelihood that evolving system dynamics will push the system across critical thresholds within a future time horizon. Modeling degradation propagation therefore requires a framework capable of capturing both interdependencies among institutional components and their dynamic evolution over time.

## 4. DCNAR as a Degradation Modeling Framework

To model degradation propagation in complex institutional systems, we employ Dynamic Causal Network Autoregression (DCNAR) [5], a trajectory-aware framework that integrates data-driven causal discovery with time-varying dynamic modeling. Unlike conventional dynamic causal models that assume the underlying causal network is known in advance, DCNAR learns this structure directly from the data and then uses it as a structural constraint for dynamic inference. The framework consists of two stages: (1) causal network discovery, which identifies directed relationships among institutional indicators, and (2) time-varying network autoregression, which models the temporal evolution of the system under this learned structure.

### 4.1. Causal Network Discovery

In the first stage, DCNAR estimates the causal reliability network among institutional indicators using a neural autoregressive causal discovery model. For each variable $x_{i,t}$, the model decomposes its dynamics into additive contributions from lagged values of all other variables,

$$x_{i,t} = \sum_{j=1}^{N} \sum_{l=1}^{L} f_{ijl}\left(x_{j,t-l}\right) + \varepsilon_{i,t}$$

Here $f_{ijl}(\cdot)$ are nonlinear neural functions that capture the influence of lagged variable $j$ on variable $i$. Aggregating these contributions yields a matrix of directed causal scores $S_{ij} = \sum_{\ell=1}^{L} \| f_{ij\ell} \|$. This matrix summarizes predictive causal dependencies among institutional indicators and defines a sparse directed network of degradation propagation. The resulting adjacency matrix $A = (a_{ij})$ represents pathways through which deterioration in one subsystem may influence others. Rather than treating this network as a final causal claim, DCNAR interprets it as a structural prior that constrains subsequent dynamic modeling. This design allows causal structure to be learned from data while maintaining interpretability and stability of the dynamic system.

### 4.2. Dynamic Autoregressive Modeling

In the second stage, the learned network is incorporated into a time-varying network autoregressive model. Let $X_t = \left(x_{1t}, x_{2t,} \dots, x_{nt}\right)$ denote the vector of institutional indicators at time $t$. The dynamic system is modeled as $X_{t+1} = f_\theta(X_t, A) + \varepsilon_t$, where $f_\theta$ is a neural autoregressive function and $A$ constrains causal interactions among variables. More formally, the DCNAR dynamic component can be expressed as a time-varying network autoregression

$$X_t = \sum_{\ell=1}^{p} (G + I)\, \Lambda_\ell(t) X_{t-\ell} + \varepsilon_t,$$

where $G$ is the learned adjacency matrix, $I$ is the identity matrix, and $\Lambda_\ell(t)$ contains time-varying influence parameters. This formulation allows causal influence among components to evolve smoothly over while remaining consistent with the learned structure.

### 4.3 Trajectory-Based Risk Prediction

Traditional survival models estimate failure risk using the current system state [6-7]. In contrast, DCNAR evaluates risk using predicted future trajectories of the institutional system. Starting from the current state $X_t$, the model recursively generates forecasts: $\hat{X}_{t+1}, \hat{X}_{t+2}, \dots, \hat{X}_{t+h}$, where $h$ is the prediction horizon. Failure risk is defined as the probability that the predicted trajectory crosses a degradation threshold within this horizon, $P\left(\min_{k \le h} \hat{x}_{t+k} \le \tau\right)$, where $\tau$ is a failure threshold associated with institutional degradation. This trajectory-based formulation captures a key feature of reliability in complex systems: failures may arise not from the current state alone but from propagation dynamics that push the system toward instability over time. By modeling the interaction network and simulating future trajectories, DCNAR enables detection of propagation-driven failures that are difficult to identify using static hazard models.

## 5. Experimental Design

### 5.1. Data

Our empirical evaluation uses a panel dataset of institutional indicators derived from the Varieties of Democracy (V-Dem) [11] project, which provides annual country–year measures of multiple dimensions of democratic governance. The dataset contains 15 variables on 139 countries observed over a relatively short time horizon, forming a panel observed annually for approximately 35 years. This structure creates many heterogeneous but relatively short time series, a setting that is common in comparative political research and presents significant challenges for dynamic causal modeling. The indicators represent multiple components of democratic systems, including electoral processes, civil liberties, institutional constraints on executive authority, and citizen participation. Together these indicators form a multivariate representation of the institutional state of each political system over time.

From a reliability perspective, these indicators function as degradation signals describing the health of institutional subsystems. Failure events are operationalized by threshold crossings in these indicators. Specifically, for each indicator we define a failure event when its value falls below the 20th percentile of its training distribution, corresponding to a degraded institutional state relative to historical observations.

This empirical setting represents a demanding test environment for dynamic reliability modeling. Short time series limit the feasibility of estimating highly parameterized time-varying models, while cross-country heterogeneity complicates the pooling of information across units. At the same time, the substantive questions motivating this study – how institutional components influence one another and how degradation propagates through the system – are inherently dynamic and causal.

### 5.2. Models

We compare three models for predicting institutional failure risk. The first baseline is a Cox proportional hazards model [7], which estimates failure risk as a function of the current system state using the proportional hazards assumption. This model represents a standard survival analysis approach in which the hazard of failure depends on the present values of institutional indicators. The second baseline is a discrete-time hazard model [12], implemented as a logistic regression estimating the probability of failure in the next period conditional on the current system state. This model relaxes the proportional hazards assumption while retaining the same static risk structure.

The third model is the proposed DCNAR trajectory risk model. Instead of estimating risk solely from the current state, the DCNAR model generates predicted trajectories of institutional indicators using the dynamic causal network autoregression. Failure risk is then defined as the probability that these predicted trajectories cross the degradation threshold within a fixed horizon. In the experiments reported here, the prediction horizon is five periods, allowing the model to capture medium-term degradation propagation.

### 5.3. Evaluation Metrics

Model performance is evaluated using several complementary metrics that capture different aspects of risk prediction. AUROC [13] (Area Under the Receiver Operating Characteristic Curve) measures the ability of a model to discriminate between failure and non-failure events. It reflects the model's ranking accuracy across all classification thresholds. AUPRC [13] (Area Under the Precision-Recall Curve) evaluates performance for rare events, emphasizing the trade-off between precision and recall when failures are relatively infrequent. Brier score [14] measures the accuracy of probabilistic predictions by computing the mean squared difference between predicted probabilities and observed outcomes. Expected Calibration Error (ECE) [15] assesses how well predicted probabilities correspond to empirical event frequencies, providing a measure of probability calibration. Together these metrics allow to evaluate both the discrimination ability and probabilistic reliability of each model in predicting institutional degradation events.

## 6. Results

Figure 2 summarizes the comparative performance of the trajectory-based DCNAR model relative to classical survival models across institutional indicators. The figure highlights two key comparisons: whether DCNAR outperforms Cox proportional hazards models and whether it outperforms discrete-time hazard models in predicting institutional failure events. Failure events are defined as indicators falling below the 20th percentile of the training distribution, and risk is evaluated over a five-period prediction horizon.

The results reveal a consistent pattern across institutional indicators. First, the DCNAR trajectory model outperforms the Cox proportional hazards model for majority of indicators, achieving higher AUROC scores than Cox. This improvement is also reflected in other metrics, including Brier score and ECE, indicating that trajectory-based modeling generally produces better calibrated probability estimates than the classical proportional hazards approach. These results suggest that dynamic modeling of institutional interactions provides a more accurate representation of failure risk than static survival models that rely solely on the current system state.

The comparison between DCNAR and the discrete-time hazard model reveals more heterogeneous results. Hazard models outperform DCNAR for several indicators whose failures appear to depend primarily on the current level of the indicator itself, such as clean elections or civil participation. In these cases, risk behaves largely as a threshold process in which the present state contains most of the relevant information for near-term failure prediction.

However, the DCNAR trajectory model clearly dominates the hazard model for a subset of institutional indicators, including local government effectiveness, direct vote mechanisms, regional government structures, and elected officials, producing substantial gains in AUROC relative to both models. The largest improvements occur for elected officials and direct vote indicators, where trajectory-based modeling captures dynamic propagation effects that static hazard models cannot represent.

This pattern suggests that institutional failures fall into two broad categories. For some indicators, failure risk is primarily state-driven, meaning that the probability of degradation is largely determined by the current value of the indicator. In these cases, classical hazard models perform well. For other indicators, however, failures are propagation-driven, emerging from interactions among institutional components that gradually push the system toward a degraded state. In these cases, trajectory-aware models such as DCNAR provide superior predictive performance because they account for how degradation evolves through the institutional network over time.

Results demonstrate that trajectory-aware modeling provides a meaningful advantage for predicting failures driven by dynamic institutional interactions, while classical survival models remain competitive when failures depend primarily on the current system state. This distinction highlights the value of incorporating propagation dynamics into reliability analysis of complex institutional systems.

## 7. Discussion and Conclusion

This study applies a reliability perspective to the analysis of institutional stability in democratic systems. By treating institutional indicators as interacting subsystems and defining degradation thresholds as failure events, democratic dynamics can be analyzed as a system reliability problem. The central question examined is whether modeling degradation propagation improves failure risk prediction relative to classical models.

The results show that the trajectory-based DCNAR model consistently outperforms the Cox proportional hazards model and provides improvements over discrete-time hazard models for several institutional indicators. These results suggest that many institutional failures are not driven solely by the current state of an indicator but emerge through dynamic interactions among institutional components. For indicators where failure risk is primarily state-dependent, hazard models remain competitive, while propagation-driven failures benefit from trajectory-based modeling.

Overall, the findings indicate that reliability analysis of democratic systems should distinguish between state-driven and propagation-driven failures. Classical hazard models capture threshold-based risks effectively, whereas trajectory-aware models such as DCNAR provide advantages when failures arise through dynamic degradation across interacting subsystems. This perspective highlights the value of incorporating system dynamics and interaction structures into risk analysis of complex institutional systems.

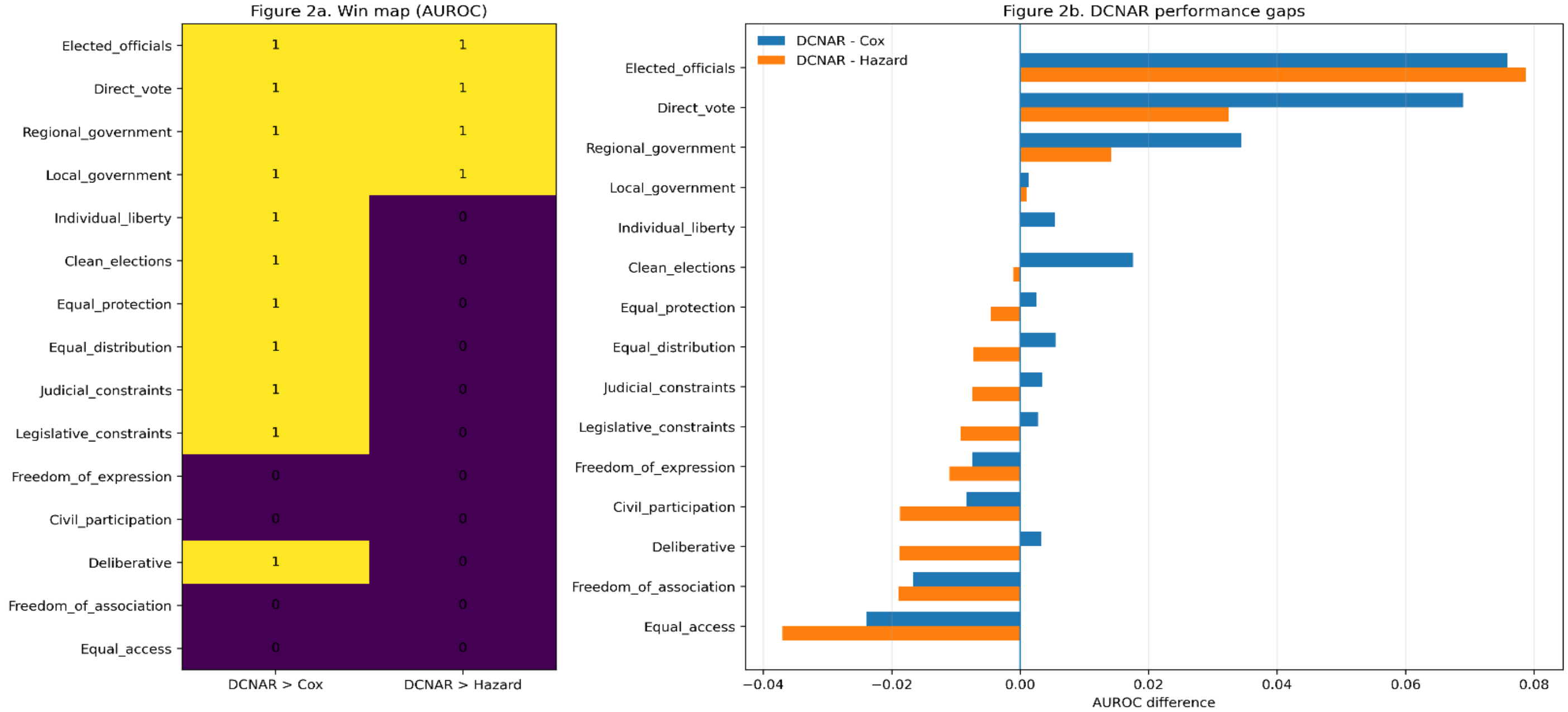

Figure 2. Comparative Performance of the Trajectory-Based DCNAR Relative to Hazard and Cox Models. "1" in Figure 2a indicates that DCNAR outperforms the comparative model. Figure 2b shows the differences in performance between DCNAR and comparative models.

## 8. Acknowledgements

This research is made possible in part by support from the Franco Institute for Liberal Arts and the Public Good, College of Arts & Letters; the Kellogg Institute for International Studies; and the Lucy Family Institute for Data & Society, University of Notre Dame. Generative AI systems (ChatGPT 5.3) were used to polish author-written text.